%% file: acl_lualatex_arxiv.tex
\documentclass[11pt]{article}

\usepackage[final]{acl}
\usepackage[table]{xcolor}
\usepackage{amsmath}
\usepackage{amssymb}
\usepackage{unicode-math}
 \usepackage{microtype}
 \usepackage{todonotes}
\usepackage{comment} 
\usepackage{booktabs}
\usepackage{graphicx}
\usepackage{array}
\usepackage{colortbl}
\usepackage{hyperref}
\usepackage{float}
\usepackage[nameinlink,capitalize,noabbrev]{cleveref}
\usepackage{enumitem}
\usepackage[english,bidi=default]{babel} % English as the main language.
\babelfont{rm}{TeXGyreTermesX} % similar to Times
\definecolor{plotblue}{HTML}{3182BD}
\definecolor{plotred}{HTML}{DE2D26}

\DeclareRobustCommand{\redcirc}{\tikz[baseline=-0.6ex]\fill[plotred] (0,0) circle (0.6ex);}
\DeclareRobustCommand{\reddiam}{\tikz[baseline=-0.6ex]\fill[plotred,rotate=45] (-0.53ex,-0.53ex) rectangle (0.53ex,0.53ex);}

\def\JH#1{{\color{blue}JH: \emph{#1}}}

\title{How to Evaluate When You Don't Know How to Evaluate?\\ Choosing Reliable Setups for Base Language Model Benchmarking}
\author{Shenbin Qian \quad David Samuel \quad Jindřich Helcl \\ 
\textbf{Andrey Kutuzov} \quad \textbf{Stephan Oepen} \\
Department of Informatics \\ 
University of Oslo \\
\{shenbinq, davisamu, jindrich, andreku, oe\}@ifi.uio.no}

\begin{document}

\maketitle
\begin{abstract}

Evaluation of large language models (LLMs) remains largely pre-scientific: it involves many design decisions, with little guidance on how to make them.
These choices can substantially affect measured performance and, consequently, decisions about subsequent pretraining.
To identify effective evaluation setups, we investigate 8 aspects, such as task formulation or the number of in-context examples.
We compare the setups using three widely-used signal criteria, repurposed as normative meta-measures: monotonicity, non-randomness, and ranking consistency of the evaluation scores across intermediate checkpoints.
In an early pretraining scenario, we evaluate these setups on 26 monolingual 2B-parameter models trained on different types of data, using 80 tasks in four languages. 
We also examine the setups on three larger LLMs ($\geq$7B) in the same four languages  and explore how the preferred setups change with model scale.
Our results yield concrete recommendations: some choices, such as the use of multiple prompts, are consistently useful, whereas others, such as task formulation, depend on model scale. 
More generally, our framework serves as a recipe for the choice of evaluation setups in new scenarios.

\end{abstract}

\section{Introduction}

\begin{figure*}[t]
\centering
\includegraphics[width=0.95\textwidth]{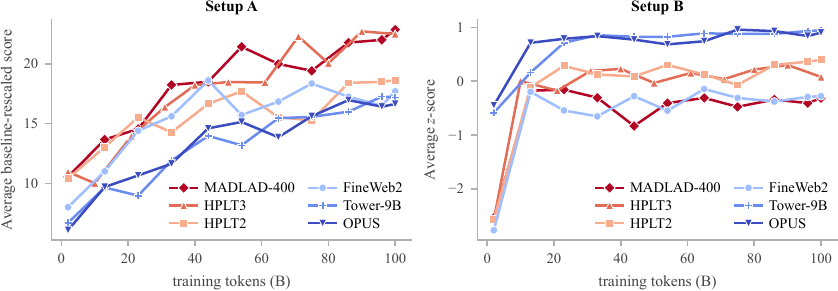}
\caption{Average score of six Norwegian 2B models, pretrained on different corpora under two evaluation setups for NorEval~\citep{mikhailov-etal-2025-noreval} classification tasks. Both use 5-shot prompting and the maximum over multiple prompts. \textbf{Setup A}: random-baseline rescaling, mean over task categories, cloze formulation (CF), no log-probability normalization, accuracy. \textbf{Setup B}: \textit{z}-score rescaling, mean over tasks, multiple-choice formulation (MCF), character normalization, answer probability. At the end of training, the two setups rank the models largely in reverse order (Kendall's $\tau = -0.6$). MADLAD-400 ranks highest under setup A but lowest under setup B, whereas the two machine-translated corpora, OPUS and Tower-9B, shift from the bottom to the top of the ranking. The choice of evaluation setup can lead to different conclusions about which training corpus is preferable.}
\label{fig:motivation}
\end{figure*}

Designing evaluation of large language models (LLMs) involves a number of decisions
about different aspects, such as the selection of benchmarks, prompting strategies, metrics, and their aggregation~\citep{liang2023tmlr-holistic,alzahrani-etal-2024-benchmarks}. Moreover, these decisions often depend on various circumstances, such as the model architecture and size, its training data, as well as the specific stage of the model training~\citep{pmlr-v267-magnusson25a}. The choices made in this process at a given stage of development, such as early pretraining, affect subsequent decisions in further model development (e.g. the training data composition for the next training phase), which in turn directly affect the final model performance.

In this meta-study, we look at such crossroads in LLM development and give advice on how to select an evaluation setup that extracts the most signal from the intermediate model results. Early pretraining will serve as our example situation: consider a model developer deciding which dataset to use to train their model. Relying on scaling laws \citep{Kaplan2020-uo}, they train a set of small proxy models on the candidate datasets and pick the dataset whose proxy performs the best.

The challenge lies in how we extract the signal from the model performance. In two case studies, we demonstrate that seemingly minor changes to the evaluation setup can lead to very different development decisions as shown in \Cref{fig:motivation}. We investigate this issue during early pretraining using intermediate checkpoints of small models, and examine its generalizability to larger model scales using three pretrained multilingual LLMs.

%Figure~\ref{fig} illustrates the sensitivity of model rankings to the evaluation setup. Under two different setups, the same six Norwegian 2B models, trained on different corpora, are ranked in largely reverse order. MADLAD-400 ranks highest under setup A but lowest under setup B, whereas the two machine-translated corpora, OPUS and Tower-9B, shift from the bottom to the top of the ranking. Consequently, the choice of evaluation setup can lead to different conclusions about which training corpus is preferable.

\paragraph{Contributions.} First, building on techniques from prior work~\citep{madaan2025quantifying, gu-etal-2025-olmes,penedo2025fineweb2}, we develop a meta-evaluation framework to identify useful evaluation setups. The framework combines three normative meta-measures, each of them quantifying one desirable property of the signal from running the evaluations: \emph{monotonicity} (performance improving with training progress), \emph{non-randomness} (clear separation from a random baseline), and \emph{ranking consistency} (stable relative rankings of models in later training stages).

Second, we conduct a systematic evaluation of eight aspects of the evaluation setup across checkpoints of 26 monolingual 2B models and 80 tasks in four languages (Norwegian, Finnish, French, and Spanish). As part of this study, we extend 14 Norwegian multiple-choice tasks in NorEval~\citep{mikhailov-etal-2025-noreval} with additional task formulations, log-likelihood normalization methods, and soft probabilistic metrics, while largely retaining existing task implementations for the other languages.

Third, we identify evaluation choices that generalize across languages and model scales, as well as those that depend on the evaluation context. Multiple prompts and 5-shot prompting provide consistently informative signals, whereas the relative effectiveness of task formulations and metrics changes with model scale: cloze formulation and answer probability are preferred for smaller models, while multiple-choice formulation and accuracy become increasingly competitive for larger models. We validate these findings on three larger multilingual LLMs. Our framework offers a systematic approach to selecting useful evaluation setups, enabling more reliable decisions during pretraining.

\section{Related Work}

\paragraph{Meta-measures and task selection.} To our best knowledge, the notion of ``meta-measuring'' LLM evaluation was first used informally to select tasks for the DCLM and FineWeb suites \citep{li2024datacomplm, 3737916.3738886}. FineTasks \citep{kydlicek2024finetasksmultilingualtasks} subsequently formalized four meta-measures: \textit{monotonicity}, independently investigated around the same time by \citet{madaan2025quantifying}, \textit{non-randomness}, \textit{ranking consistency} and \textit{signal-to-noise ratio}. We adopt three of them, following their definitions in FineWeb2 \citep[$\S$A.5.1]{penedo2025fineweb2}. These measures have been applied in several task-selection studies \citep{oepen-etal-2026-hplt, kytoniemi2025finbenchv2, alrashed2026mixminhashmatchcrosssource, yagoubi2025neurips2025e2lmcompetition}. Unlike our work, they focus on selecting \textit{which tasks} to include rather than systematically selecting the evaluation setup.

\paragraph{Evaluation setups and benchmarks.} OLMES \citep{gu-etal-2025-olmes} standardizes formulation, normalization and few-shot examples for English multiple-choice tasks, but chooses them by convention and by the accuracy they yield, not by any explicit signal-quality criteria. \citet{madaan2025quantifying} and \citet{heineman2025signal} investigate evaluation reliability, showing that continuous metrics, cloze formulation and checkpoint merging reduce noise, but primarily focus on individual evaluation setup choices. In contrast, we compare complete setups, jointly over eight aspects and three meta-measures.

Such a comparison needs benchmarks that expose these choices, but most existing benchmarks provide limited variation in prompts, formulations, metrics, or decoding strategies. NorEval \citep{mikhailov-etal-2025-noreval} provides multiple prompts per task, and FIN-bench-v2 \citep{kytoniemi2025finbenchv2} in addition contains cloze and multiple-choice formulations with five prompts each. We extend NorEval with additional formulations, log-probability normalizations, probabilistic metrics and sampling ($\S$\ref{benchmarks}) to study all eight aspects.

% BetterBench: https://arxiv.org/abs/2411.12990 --> skipping this for now, the focus seems to be more on "best practices" in benchmarking

% \JH{maybe something called DataDecide? My LLMs keep talking about this} --> note sure why, it's about selecting pretraining data. There's "How can we make evaluation benchmarks more predictable?" section, but it's AllenAI again and the recommendations are based on their previous work

% \cite{signal_noise2025}

\section{Experimental Setup}

This section introduces the aspects and options of the evaluation setup ($\S$\ref{eval_setup}), the meta-measures for evaluating the choice of evaluation setup ($\S$\ref{meta_metrics}), and the benchmarks ($\S$\ref{benchmarks}) and models ($\S$\ref{models}) we run. Inference details are in Appendix~\ref{app:inference}.

\subsection{Evaluation Setup} \label{eval_setup}

We study eight aspects of the evaluation setup. As classification (multiple-choice questions) and generation tasks are scored with different metrics, we evaluate and compare them separately. Four aspects apply to both task types, three to classification tasks only, and one to generative tasks only. 

\subsubsection{Task-Agnostic Aspects}

\paragraph{Number of shots.}
It can be difficult for a base model to infer the intended task from a short prompt alone (\emph{0-shot} setting). This prompt ambiguity can be resolved by including some solved examples of the same task in the context (\emph{5-shot} evaluation in our case), following the in-context learning approach from \citet{NEURIPS2020_1457c0d6}.

\paragraph{Prompts.}
The issue with prompt ambiguity might be worse when a task relies only on a single prompt template \citep{mizrahi-etal-2024-state}. 
%Most tasks come with several prompt templates (up to six) that phrase the same instruction differently. 
We compare three options: \emph{single}, which evaluates each task with its first (or only) template, and \emph{multi-mean} and \emph{multi-max}, which evaluate each task using all of its templates and combine the per-prompt scores by their mean or maximum.

\paragraph{Score rescaling.}
Tasks differ in their chance level %(e.g., 50\% for 2 vs.\ 20\% for 5 answer options) 
and in the range their metric attains. Therefore, simple averaging raw scores may let some tasks dominate the aggregate. We compare four ways of putting the task scores $y$ on a common footing before aggregating across tasks:
\begin{align}
    r_{\text{baseline}}(y) &= \max\!\left(0,\ \frac{y - b_q}{1 - b_q}\right), \\
    r_{\text{\textit{z}-score}}(y) &= \frac{y - \mathrm{mean}_q}{\mathrm{sd}_q}, \\
    r_{\text{min-max}}(y) &= \frac{y - \min_q}{\max_q - \min_q}, \\
    r_{\text{none}}(y) &= y.
\end{align}
Here, all scores $y$ are on a 0--1 scale, where 1 is the best possible score. $b_q$ is the random baseline of task $q$, and $\mathrm{mean}_q$, $\mathrm{sd}_q$, $\min_q$ and $\max_q$ are computed over the scores of task $q$ for all models and their checkpoints. Baseline rescaling maps chance level to 0 and a perfect score to 1 \citep{srivastava2023beyond}. %For metrics whose chance level is 0, it leaves the scores unchanged and thus is equal to \emph{none}. 
\textit{Z}-score and min-max rescaling align both the location and the spread of the tasks, but no longer have absolute interpretation \citep{10.1016/j.patcog.2005.01.012}. %at the cost of scores that are relative to the compared models and that no longer have an absolute interpretation. \JH{maybe delete from ", at the cost" or reduce to "but no longer have..."}%Baseline rescaling is applied to each score before aggregation over prompts; z-score and min-max rescaling are applied to the task scores after aggregation over prompts and formulations.

\paragraph{Cross-task aggregation.}
The task scores are combined into one score per checkpoint either by the \emph{task-level mean} over all tasks, or by the \emph{category mean}, which first averages the tasks within each task category (e.g., reading comprehension, machine translation) and then averages those scores.
This equalizes the contribution of each category to the score, regardless of how many tasks it contains.

\subsubsection{Classification Task Aspects}

For a multiple-choice question with context $x$ and answers $a_1, \dots, a_n$ (usually spanning more tokens $a_i^j$), of which $a_{i^*}$ is correct, the model scores each answer by its conditional log-likelihood:

\begin{equation}
    \ell_i = \sum_{a_{i}^{j} \in a_i} \log P_\theta(a_i^j|x, a_{i}^{<j}).
\end{equation}
The three aspects below concern how the question is presented to the model (formulation), how the option scores $\ell_i$ are normalized (log-probability normalization), and how the normalized scores are turned into a task score (metric).

\paragraph{Formulation.}
We consider three task formulations \citep{robinson2023leveraging, gu-etal-2025-olmes}, %illustrated using the Norwegian reading comprehension task NorBelebele~\citep{bandarkar-etal-2024-belebele,maehlum-etal-2025-improved} in \Cref{tab:formulation_example}.
illustrated in \Cref{tab:formulation_example}. First, in \emph{cloze formulation} (CF), the context contains only the passage and the question. The options $a_i$ are the full answer texts. The model thus scores each answer text as a continuation without seeing the alternatives. Second, in \emph{multiple-choice formulation} (MCF), the answer options are listed in the context with letter labels (e.g., A, B, C, D), and the options $a_i$ are the letters themselves. The model has to see all alternatives and map its choice to a label, which requires in-context understanding of the multiple-choice format. Third, the \emph{hybrid} formulation lists the answer options in the context, as in MCF, but without labels, and the options $a_i$ are the full answer texts, as in CF. The model sees all alternatives but does not need to produce a symbolic label.

\begin{table*}[t]
\centering
\small
\resizebox{0.9\linewidth}{!}{%
\begin{tabular}{l*{3}{>{\ttfamily\raggedright\arraybackslash}p{0.26\textwidth}}}
\toprule
 & \normalfont\textbf{CF} & \normalfont\textbf{MCF} & \normalfont\textbf{Hybrid} \\
\midrule
Context $x$ &
Question: \{question\}\newline
Answer: &
Question: \{question\}\newline
A: \{answer 1\} \dots\newline
D: \{answer 4\}\newline
Answer: &
Question: \{question\}\newline
- \{answer 1\} \dots\newline
- \{answer 4\}\newline
Answer: \\
\midrule
Options $a_i$ &
\{answer 1\}, \dots, \{answer 4\} &
A, B, C, D &
\{answer 1\}, \dots, \{answer 4\} \\
\bottomrule
\end{tabular}%
}
\caption{Example prompt templates and answer options for the three task formulations of a multiple-choice task.}
\label{tab:formulation_example}
\end{table*}

\paragraph{Log-probability normalization.} 
% While in theory, the raw log-likelihood is the only correct calculation, it suffers from multiple problems in practice when used to rank different options with an imperfect language model. It can favor short options, 
In practice, the raw log-likelihood $\ell_i$ can favor short options, since every additional token lowers it. It also reflects how likely an option is regardless of the question \citep{schaeffer2025why}. We compare three normalizations:
\begin{align}
    s_i^{\text{none}} &= \ell_i, \\
    s_i^{\text{char}} &= \ell_i \,/\, |a_i|_{\text{char}}, \\
    s_i^{\text{PMI}} &= \ell_i - \log P_\theta(a_i),
\end{align}
where $|a_i|_{\text{char}}$ is the number of characters of $a_i$, and $P_\theta(a_i)$ is the prior probability of $a_i$. Character-length normalization removes the length bias, and point-wise mutual information (PMI) removes the bias of the model for options that are likely, regardless of the question.

\paragraph{Metrics.} \label{metrics} With the normalized scores $s_1, \dots, s_n$ of the answers, we define three metrics, each averaged over the task examples:
\begin{align}
    \mathrm{Acc(s_i)} &= \mathbb{1}\big[\arg\max_i s_i = i^*\big], \\
    \mathrm{CondProb(s_i)} &= \frac{\exp(s_{i^*})}{\sum_{i=1}^{n} \exp(s_i)}, \\
    \mathrm{AnsProb(s_i)} &= \exp(s_{i^*}).
\end{align}
These metrics are based on the earlier work by \citet{schaeffer2025why}. \emph{Accuracy} (Acc) is discrete: each question contributes 0  or 1, so a model whose probability for the correct answer rises steadily during training gains nothing until that answer overtakes the others. The two probability-based metrics are continuous and can therefore reveal such gradual progress in early pretraining. The \emph{conditional answer probability} (CondProb) is the probability mass that a softmax over the normalized scores assigns to the correct option, i.e., the probability of the correct option conditional on the set of answer options. The \emph{answer probability} (AnsProb) is the probability of the correct option itself, without comparison to the alternatives.\footnote{AnsProb and CondProb are true probabilities of the model only with $s^{\text{none}}$; with $s^{\text{PMI}}$,
AnsProb can exceed~1.}
% Both character-length normalization and PMI can be applied to the three metrics to get a full set of 9 metrics for classification tasks.

%With character normalization, the softmax is taken over per-character scores, which flattens the distribution. The \emph{answer probability} (AnsProb) is the probability of the correct option itself, without comparison to the alternatives: $P_\theta(a_{i^*} \mid x)$ without normalization, and the per-character (geometric-mean) probability $P_\theta(a_{i^*} \mid x)^{1/|a_{i^*}|_{\text{char}}}$ with character normalization. With PMI, $\exp(s_{i^*}^{\text{PMI}}) = P_\theta(a_{i^*} \mid x) / P_\theta(a_{i^*})$ is a ratio rather than a probability and spans many orders of magnitude, so we use the log-ratio $s_{i^*}^{\text{PMI}}$ instead. Accuracy and the conditional answer probability have the random baseline of the task as their chance level (e.g., $1/n$ for $n$ options with one correct answer), whereas the chance level of the answer probability is 0. All nine combinations of metric and normalization are available for Norwegian ($\S$\ref{benchmarks}). % For Finnish and Spanish, we have accuracy with all three normalizations (with PMI for only 5 of the 11 Spanish tasks) and the character-normalized answer probability. For French, we have accuracy with and without character normalization, and the character-normalized answer probability.

\subsubsection{Generation Task Aspects}

\paragraph{Decoding.}
We narrow our selection of decoding strategies for generation tasks to two options. With \emph{greedy decoding}, the model generates the most likely token at every step, which gives a single deterministic output per question. With \emph{random sampling}, tokens are sampled from the full model distribution (temperature 1, no top-$k$ or top-$p$ truncation), estimating the expected performance of the evaluated model \citep{miller2024addingerrorbarsevals}. %This comparison is only available for the Norwegian generation tasks.

\subsection{Meta-Measures}
\label{meta_metrics}
%Note: thresholds of training steps for meta metrics

We use three \emph{meta-measures} to assess how informative and diagnostic an evaluation setup is for tracking model development during pretraining. Whereas an ordinary evaluation metric assigns a score to a model checkpoint on a task, a meta-measure operates on a collection of such scores across checkpoints and models. We define the three meta-measures formally in the following subsections. Appendix~\ref{app:methods} describes how we compute them for all setups and languages.

\paragraph{Notation.}
Let $f \in \mathcal{F}$ index a model, $t = 1, \dots, T_f$ its checkpoints, and $p_{f,t}$ the number of training tokens at checkpoint $t$, so that $p_{f,T_f}$ is the token budget of a model $f$. An evaluation setup $e$ assigns one option to each aspect in $\S$\ref{eval_setup}. All setup analyses are carried out separately for each language and task type. Under setup $e$, each checkpoint receives a single score $s_{f,t}$, which is computed from task scores with the metric (e.g., accuracy) of $e$, aggregated over prompts, rescaled and aggregated over tasks, as prescribed by $e$. %(baseline rescaling is applied to the score of each prompt, z-score and min-max rescaling to the task scores). 
A meta-measure maps the scores of a setup over all models and their checkpoints to a single value.\footnote{We note that $s_{f,t}$ is a function of $e$. For readability, we omit this from the notation.}

\paragraph{Monotonicity.}
A useful setup should reflect that a model improves as it is trained on more data. For each model, we compute the Spearman rank correlation $\rho$ between training progress and score over all of its checkpoints, and average it across models:
\begin{equation}
    \mathrm{Mono}(e) = \frac{1}{|\mathcal{F}|}\!\sum_{f \in \mathcal{F}} \rho\Big[\!(p_{f,t})_{t=1}^{T_f}, (s_{f,t})_{t=1}^{T_f}\!\Big]\!.\!
\end{equation}
% Models whose scores are constant across checkpoints are skipped, since $\rho$ is undefined for them. %$\mathrm{Mono}(e) \in [-1,1]$, and $1$ means that the score increases with every checkpoint for every model.

\paragraph{Non-randomness.}
A setup should yield scores that are clearly distinguishable from chance, i.e., the gap between the best model and a random model should be large relative to the noise of the scores. We therefore express the gap between the best final score and the random baseline in units of noise, following FineWeb2~\citep{penedo2025fineweb2}:
\begin{equation}
    \mathrm{NonRand}(e) = \frac{\max_{f \in \mathcal{F}} s_{f,T_f} - b_e}{\hat{\sigma}_e},
\end{equation}
where $b_e$ is the score of a random model, obtained by putting the random baseline $b_q$ of every task through the same rescaling and task aggregation as the model scores. We use the variation of each model's score at the end of training as a proxy for the noise $\hat{\sigma}_e$: the sample standard deviation over its last $K=5$ checkpoints averaged across models,
\begin{equation}
    \hat{\sigma}_e = \frac{1}{|\mathcal{F}|} \sum_{f \in \mathcal{F}} \operatorname{sd}\big( s_{f,T_f-K+1}, \dots, s_{f,T_f} \big).
\end{equation}
Since $b_e$ and $\hat{\sigma}_e$ are on the same scale as the scores, $\mathrm{NonRand}$ is unit-free and comparable between setups with different score rescaling.

\paragraph{Ranking consistency.}
A setup should rank models consistently, so that the comparison drawn at one checkpoint is not overturned at the next one. Since models may differ in their token budgets (e.g., the larger LLMs in $\S$\ref{models}), we align them by the number of training tokens, and split training into consecutive windows at half of the smallest budget and at the end of the training of each model. A window is used only if at least two models are still training in it. Following FineWeb2, we ignore the first window, where rankings are very inconsistent. 

Let $\pi_{k,1}, \dots, \pi_{k,m_k}$ be the rankings of the models at the successive checkpoints in window $k$. Ranking consistency averages Kendall's $\tau$ between adjacent rankings within each window, and then over the $N$ remaining windows:
\begin{align}
    \mathrm{RankCons}(e) &= \sum_{k=1}^{N} \sum_{i=1}^{m_k-1} \frac{\tau\big( \pi_{k,i}, \pi_{k,i+1} \big)}{N(m_k - 1)}.
\end{align}
%For the 2B models, which share one token budget, this is the later half of training. For the larger LLMs, as they differ in token budget, the windows cover 3--6T tokens with all three models and 6--8T tokens with Prelude and Apertus, comparing them at the same number of training tokens.

\paragraph{From meta-measures to setup choices.} \label{interp_method}

A setup could be selected separately based on the highest score of each meta-measure, but the three meta-measures are on different scales and may favor different setups. We therefore combine them using ranks. For each language and task type, we compute an \emph{overall score} ($0$--$1$) for each setup as the mean of its percentile ranks under the three meta-measures. We evaluate setups in three complementary ways. The \emph{best complete setup} is the setup with the highest overall score. It captures interactions among aspects but may depend on small score differences. The \emph{option ranking} scores each option by the mean overall score of all setups containing it, averaging over all combinations of the other aspects, and thus measures its average effect. The \emph{per-option win rate} compares pairs of setups that differ in only one aspect and records, for each meta-measure, how often an option wins (ties count as half). We then average across the three meta-measures. This measures how consistently an option outperforms alternatives, regardless of the magnitude of the difference. Further details are provided in Appendix~\ref{app:methods}.

\subsection{Evaluation Benchmarks} \label{benchmarks}

We use an established multi-task benchmark for each of the four languages. The benchmarks differ in which evaluation setups they support. \Cref{tab:setup_coverage} lists the number of tasks that support each aspect of the evaluation setup for each language. 

\begin{figure*}[t]
    \centering
    \includegraphics[width=0.81\linewidth]{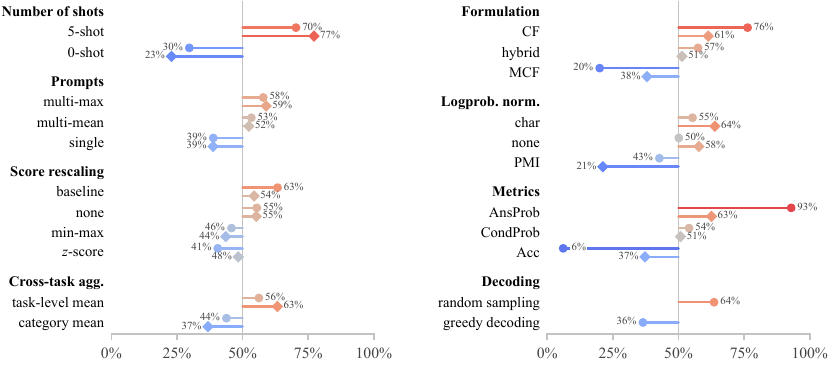}
    \caption{Per-option win rates of the 2B models (circle~\redcirc) and larger LLMs (diamond~\reddiam) pooled over languages and task types; win rates are computed per meta-measure, then averaged. Higher win rate (red) is better. The decoding aspect is only available for the 2B models.%Formulation is available only for Finnish and Norwegian; CondProb, hybrid formulation and the decoding aspect only for the 2B models in Norwegian.
    }
    \label{fig:results-summary}
\end{figure*}

\paragraph{NorEval.}
NorEval \citep{mikhailov-etal-2025-noreval} is a Norwegian benchmark of 35 tasks, 17 multiple-choice and 18 generative, covering Norwegian Bokmål, Nynorsk and Northern Sámi. Its tasks span language understanding, linguistic knowledge, world knowledge and reasoning, generation and summarization, and machine translation. Most of them come with four to six human-written prompts. Since the original benchmark does not support several of the setups studied in this work, we extend it in three ways:
\vspace{-0.5em}
\begin{itemize}
[align=parleft,left=0.5em..1.5em]
\setlength\itemsep{0.0em}
    \item \textbf{Task formulations:} every multiple-choice task with a set of answer options (14 out of 17)\footnote{Pair-wise comparison tasks such as NoCoLA \citep{jentoft-samuel-2023-nocola} are excluded.} is available in CF, MCF and hybrid formulation, for all of its prompts.
    \item \textbf{Log-probability normalization and metrics:} in addition to the raw accuracy and character-length normalized accuracy, we compute CondProb and AnsProb and their character- and PMI-normalized counterparts as additional metrics.
    \item \textbf{Random sampling:} generative tasks draw $K$ samples per question with temperature one, and report the mean score over samples. 
\end{itemize}
Together with the multiple prompts, our extension makes Norwegian the only language for which all eight aspects can be studied (\Cref{tab:setup_coverage}).

\paragraph{FIN-bench-v2.}
For Finnish, we use FIN-bench-v2 \citep{kytoniemi2025finbenchv2}, which combines Finnish versions of established English benchmarks %, such as ARC-Challenge \citep{allenai:arc}, Belebele \citep{bandarkar-etal-2024-belebele}, GoldenSwag \citep{chizhov2025hellaswagvaliditycommonsensereasoning}, TruthfulQA \citep{lin-etal-2022-truthfulqa}
with native-authored tasks in Finnish. We use its 14 tasks, 12 multiple-choice and 2 generative, each with five prompts. Ten of the multiple-choice tasks come in both CF and MCF formulation, which allows us to compare these two formulations.

\paragraph{FrenchBench.}
For French, we use FrenchBench \citep{faysse2024croissantllm}, and the French portions of Global-MMLU \citep{singh-etal-2025-global} and INCLUDE \citep{Romanou2024-ab}, resulting in 16 tasks (11 multiple-choice, 5 generative). Nine of the tasks are evaluated with the multi-prompt version from HPLT-E \citep{oepen-etal-2026-hplt}. The remaining seven have the single prompt of its original implementation. FrenchBench has no formulation variants and does not provide PMI-normalized accuracy, so French is not included in the comparisons of formulation and log-probability normalization.

\paragraph{SpanishBench.}
For Spanish, we use SpanishBench from IberoBench \citep{baucells-etal-2025-iberobench} in its multi-prompt version from HPLT-E. It comprises 16 tasks (11 multiple-choice, 5 generative), each with three prompts. There are no formulation variants for Spanish. PMI-normalized accuracy is available for only 5 of the multiple-choice tasks.

% For Finnish, French and Spanish, generative tasks use greedy decoding only, so the comparison of decoding strategies is restricted to Norwegian.

% \begin{figure}[t]
%     \centering
%     \includegraphics[width=1\linewidth]{latex/draft_figures/winrates_lollipop_scale_1col.pdf}
%     \caption{Per-option win rates of the 2B models (circle~\redcirc) and larger LLMs (diamond~\reddiam) pooled over languages and task types; win rates are computed per meta-measure, then averaged. Formulation is available only for Finnish and Norwegian; CondProb, hybrid formulation and the decoding aspect only for the 2B models in Norwegian.}
%     \label{fig:results-summary}
% \end{figure}

\subsection{Models} \label{models}

\paragraph{Early pretraining models.}
Our main experiments use 26 monolingual 2B-parameter models that share the same architecture and training setup and differ only in their pretraining data, each trained on 100B tokens of Norwegian, Finnish, French or Spanish. Four models are trained on native web corpora, FineWeb2, HPLT 2.0, HPLT 3.0 and MADLAD-400 \citep{oepen-etal-2026-hplt}, and three on synthetic data machine-translated from English with Tower-9B, Tower-72B and OPUS-MT, respectively \citep{Idahl2026-zf}. Tower-72B is available for Finnish and Spanish only, which gives seven models for these two languages and six for French and Norwegian. We evaluate 11 checkpoints of each model, from 2B to 100B tokens.

\paragraph{Larger-scale LLMs.}
To examine how the preferred setups change with model scale, we evaluate the intermediate checkpoints of three open-source larger multilingual LLMs that differ in their training budgets: 1) OLMo~3 7B \citep{olmo2025olmo3}, with 10 checkpoints of its pretraining stage up to 6T tokens; 2) Apertus 8B \citep{hernandez-cano-etal-2026-apertus}, with 14 checkpoints up to 15T tokens; 3) Prelude 9B,\footnote{\url{https://huggingface.co/openeurollm/prelude}} with 45 checkpoints up to 8T tokens.%, as its training is still ongoing.

\section{Results}
\label{results}

We first present the early pretraining models ($\S$\ref{2b_results}) and then compare them with the larger LLMs ($\S$\ref{larger_scale}). The win rates of both model scales are summarized in \Cref{fig:results-summary}. Language-specific results of the three setup choice methods are in Appendix~\ref{app:language_results}.

\subsection{Early Pretraining Models} \label{2b_results}

Based on the agreement of the three methods -- best complete setup, option ranking and per-option win rates -- we group the evaluation aspects into a tiered recipe: high-impact choices, low-impact choices and context-dependent choices in \Cref{tab:agreement}.

\paragraph{High-impact choices.}
Four choices are consistently selected by all three methods and in nearly every language for the 2B models. These choices can yield drastically different conclusions about pretraining as shown in \Cref{fig:motivation}.

The \emph{answer probability} is by far the most informative metric. It wins 93\% of its pairwise comparisons across all meta-measures. Accuracy is the least informative metric because its discrete scores increase variance without giving enough signal.

\emph{Character normalization} wins in every language except Spanish, although its win-rate advantage is small (55\%) and arises primarily from non-randomness. PMI has the highest non-randomness (66\%) but the lowest monotonicity (18\%).

\emph{5-shot} prompting outperforms 0-shot (win rate 70\%), most clearly in monotonicity (84\%).

Among the formulations, \emph{CF} is selected by all methods (win rate 76\%), whereas MCF is the least informative, consistent with the observation that small models have not yet learned the multiple-choice format \citep{gu-etal-2025-olmes}.

% \vspace{-0.5em}
% \begin{itemize}
% [align=parleft,left=0.5em..1.5em]
% \setlength\itemsep{0.0em}
%     \item The \emph{answer probability} is by far the most informative metric. It wins 93\% of its pairwise comparisons across the three meta-measures. Accuracy is the least informative metric because its discrete scores increase variance while not giving enough signal.
%     \item \emph{Character normalization} is preferred in every language except Spanish, although its win-rate advantage is small (55\%) and arises primarily from non-randomness. PMI achieves the highest non-randomness (66\%) but the lowest monotonicity (18\%).
%     \item \emph{5-shot} prompting outperforms 0-shot (win rate 70\%), with the clearest advantage in monotonicity (84\%).
%     \item Among the formulations, \emph{CF} is selected by all methods (win rate 76\%), whereas MCF is the least informative, consistent with the observation that small models have not yet learned the multiple-choice format \citep{gu-etal-2025-olmes}.
% \end{itemize}

\paragraph{Low-impact choices.} There are two aspects that do not show a clear preference.

Using \textit{multiple prompts} is preferable to using a single prompt, but the choice between aggregating their scores by the maximum or the mean has little effect. The two aggregation strategies differ primarily in the meta-measure they favor: maximum aggregation yields higher non-randomness, whereas mean aggregation yields higher monotonicity (Appendix~\ref{app:disagreement} for disagreement analysis). %The three methods do not select the same prompt option in any language, and for Spanish, two methods prefer a single prompt.

Similarly, averaging over all tasks and averaging over task-category means yield comparable results, with win rates of 56\% and 44\%, respectively. Only French favors the category mean for all three methods, whereas the other languages differ in choices.

\begin{table}[t]
\centering
\small
\setlength{\tabcolsep}{4pt}
\resizebox{0.9\linewidth}{!}{%
\begin{tabular}{@{}llcc@{}}
\toprule
\textbf{Tier} & \textbf{Aspect (top option)} & \textbf{2B} & \textbf{Larger} \\
\midrule
High-impact  & Metric (AnsProb)       & 4/4 & 3/4 \\
             & Logprob. Norm.\ (char)          & 3/4 & 3/4 \\
             & Shots (5-shot)         & 4/4 & 4/4 \\
             & Formul.\ (CF)          & 2/2 & 1/2 \\
\midrule
Low-impact   & Prompts (multi-max)    & 0/4 & 1/4 \\
             & Task agg.\ (task-level mean)      & 2/4 & 3/4 \\
\midrule
Context-dep. & Rescaling (baseline / none) & 0/4 & 0/4 \\
             & Decoding (sampling)    & 0/1 & --  \\
\bottomrule
\end{tabular}%
}
\caption{Number of languages in which all three setup choice methods select the option with the highest pooled win rate (\Cref{fig:results-summary}). For rescaling, the top option is baseline for the 2B models and none for the larger LLMs. Formulation: Finnish and Norwegian; decoding: Norwegian, 2B models only.}
\label{tab:agreement}
\end{table}

% \vspace{-0.5em}
% \begin{itemize}
% [align=parleft,left=0.5em..1.5em]
% \setlength\itemsep{0.0em}
%     \item Using \textit{multiple prompts} is preferable to using a single prompt, but the choice between aggregating their scores by the maximum or the mean has little effect. The two aggregation strategies differ primarily in the meta-measure they favor: maximum aggregation yields higher non-randomness, whereas mean aggregation yields higher monotonicity (see Appendix~\ref{app:disagreement} for the disagreement analysis). %The three methods do not select the same prompt option in any language, and for Spanish, two methods prefer a single prompt.
%     \item Similarly, averaging over all tasks or over task-category means yields nearly identical option-ranking scores (.51 vs.\ .50). Only French favors the category mean, whereas the other languages favor the unweighted mean over tasks.
% \end{itemize}

\paragraph{Context-dependent choices.} The most disagreement is in preferences of the following two aspects.

Score rescaling exhibits the greatest disagreement among the three methods. In no language do all three methods select \textit{baseline rescaling}, and their preferred options are distributed across baseline, min-max and no rescaling. This disagreement is driven by an interaction with the metric. For answer probability, whose chance level is 0, all rescaling methods yield nearly identical mean overall scores, and the best complete setups select min-max rescaling by a small margin. By contrast, for accuracy, baseline rescaling and no rescaling clearly outperform min-max and \textit{z}-score rescaling. 

For the Norwegian generative tasks, \textit{sampling} performs better on average (option-ranking score 0.56 vs.\ 0.45 in \Cref{tab:ranking_score}; win rate 64\%), although the best complete setup uses greedy decoding.

% \vspace{-0.5em}
% \begin{itemize}
% [align=parleft,left=0.5em..1.5em]
% \setlength\itemsep{0.0em}
%     \item Score rescaling exhibits the greatest disagreement among the three methods. In no language do all three methods select baseline rescaling, and their preferred options are distributed across baseline, min-max and no rescaling. This disagreement is driven by an interaction with the metric. For answer probability, whose chance level is 0, all rescaling methods yield nearly identical mean overall scores (.76--.77), and the best complete setups select min-max rescaling by a small margin. By contrast, for accuracy, baseline rescaling and no rescaling (.42 and .41) clearly outperform min-max and \textit{z}-score rescaling (.35 and .33). For the Norwegian generative tasks, sampling performs better on average (option-ranking score .56 vs.\ .45; win rate .64), although the best complete setup uses greedy decoding.
% \end{itemize}

\paragraph{The tiered recipe versus the best complete setup.}
Combining the highest-ranked option for each aspect according to the option-ranking analysis in \Cref{tab:ranking_score} (5-shot, multi-max, baseline rescaling, mean over tasks, CF, answer probability, and character normalization) yields a setup that ranks 6th among the 264 setups shared across all languages, with an overall score of 0.93, compared with 0.95 for the best complete setup. The latter differs only in its use of min-max rescaling. Thus, selecting options independently by aspect incurs only a small loss relative to jointly optimizing the complete setup, and all of the top five complete setups agree on the \textit{high-impact choices}.

\subsection{Larger-Scale LLMs}
\label{larger_scale}

We apply our approach to the three larger LLMs ($\S$\ref{models}) to assess its robustness across model scales and training stages. Specifically, we examine whether the identified recipe captures properties of the evaluation setups rather than characteristics specific to the 2B models: high-impact choices are expected to remain stable across scales, whereas context-dependent choices may vary. We also assess which evaluation setups continue to provide informative signals later in pretraining, when model performance is well above chance. \Cref{fig:results-summary} presents the win rates of smaller vs. larger models.

\paragraph{Findings that remain stable across scale.}
For the larger LLMs, all three methods continue to select answer probability with character normalization, 5-shot prompting and aggregation by the mean over all tasks, consistent with the results for the 2B models. Using multiple prompts also remains preferable to using a single prompt, with the single-prompt option attaining a win rate of 35\% for the larger LLMs, compared with 37\% for the 2B models. The three methods agree to a similar extent for the larger LLMs as shown in \Cref{tab:agreement}. The exceptions are formulation, where all three methods select hybrid, and the metric, where the win rate favors accuracy for Norwegian (\Cref{tab:picks_classification}). These results indicate that the high-impact choices identified for the 2B models are not specific to that model scale, providing further support for the use of the meta-measures to select evaluation setups.

\paragraph{Scale-dependent changes.}

The main differences arise among choices whose impacts are high or context dependent. These changes are consistent with known effects of model scale. Accuracy becomes much more informative (win rate 6\% $\rightarrow$ 37\%), primarily due to improved monotonicity (9\% $\rightarrow$ 53\%), suggesting that it provides a more reliable signal of training progress once models perform well above chance. Although answer probability remains the strongest metric, its advantage decreases (93\% $\rightarrow$ 63\%), both with monotonicity (92\% $\rightarrow$ 44\%) and ranking consistency (92\% $\rightarrow$ 52\%).

The advantage of CF over other task formulations diminishes for the larger LLMs. Although CF remains the preferred formulation overall, its win rate decreases (76\% $\rightarrow$ 61\%), whereas MCF improves (20\% $\rightarrow$ 38\%), especially in ranking consistency (33\% $\rightarrow$ 71\%) and monotonicity (2\% $\rightarrow$ 30\%). For Norwegian, the option ranking and the win rate prefer the hybrid formulation (\Cref{tab:picks_classification}). PMI normalization shows a substantial decline in non-randomness (66\% $\rightarrow$ 17\%), indicating that it becomes less effective, relative to other options, at distinguishing model performance from chance. Preferences among rescaling methods also become less pronounced. \textit{Z}-score rescaling gains (39\% $\rightarrow$ 49\%) and is used in the best complete setup, whereas option ranking and win rate slightly favor no rescaling (56\%) over baseline rescaling (54\%).

At the level of complete setups, the overall scores for the 2B and larger models are only moderately correlated (Spearman's $\rho$ = 0.66 for Finnish, 0.41 for Spanish, 0.32 for French, and 0.30 for Norwegian), with high agreement among the highest-ranked setups and substantial reordering in the middle of the ranking.

\paragraph{Implications.}
The scale-robust components of the recipe can be applied throughout pretraining, whereas context-dependent choices should be reassessed as model scale increases. During later stages of pretraining, accuracy and the MCF formulation become considerably more informative, making them increasingly viable alternatives to answer probability and CF, respectively. However, because these results are based on only three models trained with different compute budgets, estimates of ranking consistency rely on relatively few comparisons. We therefore interpret the observed scale-dependent differences as suggestive rather than conclusive.

\section{Conclusion}

We studied how the choices in an evaluation setup affect the signal that benchmark results provide during pretraining. We scored every combination of eight evaluation aspects on three meta-measures (monotonicity, non-randomness and ranking consistency), using 26 monolingual 2B-parameter models in four languages and validating on three larger multilingual LLMs.

Our results support a tiered recipe rather than a single optimal evaluation setup. Several choices consistently yield more informative signals across model scales: answer probability with character normalization outperforms alternative metrics, while 5-shot prompting and multiple prompts outperform 0-shot and single-prompt evaluation, respectively. Other choices, such as prompt and task aggregation strategies, have relatively limited effects. Some preferences can vary with model scale: answer probability and cloze formulation are preferred for the 2B models, whereas accuracy and multiple-choice (or hybrid) formulation become increasingly informative as model performance grows. These findings suggest prioritizing robust choices while reassessing context-dependent ones as training progresses. More broadly, our framework provides a systematic approach to assessing the reliability of evaluation setups before using their results to guide model development.

\section*{Limitations}

There is no ground truth for which evaluation setup is best during pretraining. We therefore use three meta-measures as a proxy for a useful indicator. These properties are desirable, but they do not guarantee that a setup captures everything that matters for pretraining decisions. For instance, a setup can score high on all three meta-measures while measuring a narrow capability, or while being insensitive to differences in data that matter downstream. Our recipe should therefore be read as a selection of setups that give informative and diagnostic signals, not of setups that are necessarily the most valid measures of model quality.

We combine the three meta-measures with equal weight, as the mean of their percentile ranks. Other combinations, such as different weights, requiring a minimum value on each meta-measure, or Pareto-optimal setups, could lead to different choices, in particular where the meta-measures disagree. Appendix~\ref{app:disagreement} discusses such a case in more detail. Other meta-measures have been proposed, such as the signal-to-noise ratio of \citet{signal_noise2025}. We did not include them, since they are defined and estimated in different ways (e.g., from seed runs or variation between nearby checkpoints) that are not directly comparable with our setting.

Our study covers four European languages, each with its own benchmark, and the available options differ between them. Formulation variants exist only for Finnish and Norwegian. The conditional answer probability and generative tasks are compared for Norwegian only. Some findings may therefore reflect properties of individual benchmarks rather than of the evaluation setups. The larger-scale validation relies on three models with different architectures, training data and token budgets. With three models, ranking consistency rests on few comparisons, so we consider the differences between the 2B and the larger models indicative rather than conclusive.

\section*{Ethical Considerations}

This research relies exclusively on publicly accessible datasets, with all data utilization adhering to the license agreements specified by~\citet{mikhailov-etal-2025-noreval,kytoniemi2025finbenchv2,faysse2024croissantllm,singh-etal-2025-global,Romanou2024-ab,baucells-etal-2025-iberobench} and~\citet{oepen-etal-2026-hplt}. It is presumed that these datasets contain no sensitive or personally identifiable information. Consequently, their application in this study is deemed to have no significant ethical risks. 

All original ideas, analyses, and content were created by the authors. AI tools were used only as supportive aids for improving writing quality and assisting with coding tasks. The authors retain full responsibility for the intellectual content, analyses, and conclusions presented in this work.

\section*{Acknowledgments}

This research was supported by the OpenEuroLLM project, co-funded by the Digital Europe Programme under grant agreement No. 101195233. This research was also supported by the European Union's Horizon Europe research and innovation programme under the Marie Skłodowska-Curie grant agreement No. 101126636. We acknowledge Sigma2, Norway for awarding this project access to the LUMI supercomputer, owned by the EuroHPC Joint Undertaking, hosted by CSC (Finland) and the LUMI consortium.

\bibliography{custom}

\appendix
\counterwithin{figure}{section}
\setcounter{figure}{0} % Reset figure counter
\counterwithin{table}{section}
\setcounter{table}{0} % Reset figure counter

\newpage

\section{Appendix: Inference Details}
\label{app:inference}

We obtain the raw results using the LM Evaluation Harness \citep{eval-harness} with vLLM~\citep{kwon2023vllm} as the backend. Each evaluation job runs one model checkpoint on a single GPU without tensor parallelism. The 2B models are loaded in their default precision and the larger LLMs in bfloat16, using the intermediate checkpoints released by their developers. For multiple-choice tasks, the harness computes the log-likelihood of every answer option given the context and, for PMI, given an empty context. The metrics in \S\ref{metrics} are derived from these log-likelihoods. Generative tasks are decoded greedily or by sampling (\S\ref{eval_setup}). The few-shot examples are drawn with a fixed seed, so that all checkpoints of all models see the same examples. The harness writes one result file per model checkpoint, task, prompt, formulation and number of shots, from which we extract the metric values.

All evaluations were run on NVIDIA GH200 GPUs (120\,GB). Evaluating the 2B models took roughly 11,000 GPU hours, most of them for the Norwegian benchmark with its many prompts, formulations and decoding strategies. Evaluating the three larger LLMs took roughly 7,000 GPU hours in total. Computing the meta-measure scores for all setups takes around 5 minutes on a CPU.

\section{Appendix: From Meta-Measure Scores to Setup Choices}
\label{app:methods}

\paragraph{Meta-measure matrices.}
For each language and task type, we enumerate all combinations of the options of the aspects (\S\ref{eval_setup}) that are available in that language. For each setup $e$, we compute the score $s_{f,t}$ of every checkpoint as described in \S\ref{meta_metrics}: we select the metric and normalization of $e$ for every task the setup covers, apply baseline rescaling to the score of each prompt, aggregate over the prompts of $e$ (the first template only, or the mean or maximum over all templates), apply \textit{z}-score or min-max rescaling to the task scores, and aggregate over tasks or task categories. From these scores, we compute the three meta-measures. This yields a matrix of meta-measure scores $M_{\mu,l}(e)$ with one entry per meta-measure $\mu$, language $l$ and setup $e$. An entry is undefined if language $l$ does not offer an option of $e$. For the 2B models, the classification matrix covers 1,488 setups and the generation matrix (Norwegian) 96 setups. 

\paragraph{Overall score.}
Within each language, we convert each meta-measure to its percentile rank among the setups of that language, with ties receiving their average rank, and define the overall score of a setup as the mean of its three percentile ranks:
\begin{equation}
    O_l(e) = \frac{1}{3} \sum_{\mu} \mathrm{pct}_l\big( M_{\mu,l}(e) \big).
\end{equation}

\paragraph{Best complete setup.}
Within a language, the best complete setup is $\arg\max_e O_l(e)$. Pooled over languages, it maximizes the mean overall score $\frac{1}{|\mathcal{L}|}\sum_{l} O_l(e)$ over the setups that every language has. For languages without an option in an aspect (e.g., French and Spanish for formulation), that aspect is ignored: we let them contribute the same score to all of its options, so the aspect is decided by the other languages.

\paragraph{Option ranking.}
For an aspect $a$ and one of its options $o$, the option ranking averages the overall score over all setups in which aspect $a$ takes option $o$, i.e., over all combinations of the other aspects:
\begin{equation}
    R_l(a, o) = \operatorname{mean}_{e:\, e_a = o} O_l(e).
\end{equation}
Pooled over languages, $R(a,o)$ is the mean of $R_l(a,o)$ over the languages $\mathcal{L}_a$ that offer an option in aspect $a$. Combining the best option of each aspect gives the setup reported as the option-ranking.

\paragraph{Per-option win rate.}
For an aspect $a$ and two of its options $o$ and $o'$, a \emph{context} $c$ is a combination of options of all other aspects, and $e_c^{o}$ and $e_c^{o'}$ are the two setups that differ only in aspect $a$. Within a language and for each meta-measure, option $o$ wins a context if it has the higher meta-measure score, with ties counting half:
\begin{equation}
    w_{\mu,l}(c; o, o') =
    \begin{cases}
        1 & \text{if } m_{c}^{o} > m_{c}^{o'}, \\
        \tfrac{1}{2} & \text{if } m_{c}^{o} = m_{c}^{o'}, \\
        0 & \text{otherwise,}
    \end{cases}
\end{equation}
with $m_{c}^{o} = M_{\mu,l}(e_c^{o})$.
The win rate of option $o$ is the share of all its comparisons with the other options of aspect $a$ that it wins:
\begin{equation}
    W_{\mu,l}(a, o) = \frac{\sum_{o' \neq o} \sum_{c \in C_l(o,o')} w_{\mu,l}(c; o, o')}{\sum_{o' \neq o} |C_l(o,o')|},
\end{equation}
where $C_l(o,o')$ is the set of contexts in which both options exist in language $l$. We average $W_{\mu,l}(a,o)$ over the languages in $\mathcal{L}_a$ and over the three meta-measures, and select the option with the highest win rate. Unlike option ranking, which uses the overall score, the win rate compares the raw meta-measure values and ignores the size of the differences.

\section{Appendix: Task Coverage of Evaluation Setups}
\label{app:setup_coverage}

\Cref{tab:setup_coverage} lists, for each aspect, the compared options and the number of tasks per language on which they are evaluated. A setup covers the tasks for which all of its options are available, so the number of tasks can differ between the options of an aspect.

\begin{table*}[t]
\centering
\small
\setlength{\tabcolsep}{4pt}
\resizebox{0.99\linewidth}{!}{%
\begin{tabular}{@{}llcccc@{}}
\toprule
\textbf{Aspect} & \textbf{Options} & \textbf{Finnish} & \textbf{French} & \textbf{Norwegian} & \textbf{Spanish} \\
\midrule
Number of shots & 0-shot, 5-shot & 10 & 11 & 14 & 11 \\
Prompts & single, multi-mean, multi-max & 10 & 11 & 14  & 11  \\
Rescaling & baseline, \textit{z}-score, min-max, none & 10 & 11 & 14 & 11 \\
Task aggregation & task-level mean, category mean & 10 & 11 & 14 & 11 \\
Formulation & CF / MCF / hybrid & 10 / 10 / -- & -- & 14 / 14 / 14 & -- \\
Accuracy & none / char / PMI & 10 / 10 / 10 & 11 / 11 / -- & 14 / 14 / 14 & 11 / 11 / 5 \\
CondProb & none / char / PMI & -- & -- & 14 / 14 / 14 & -- \\
AnsProb & none / char / PMI & -- / 10 / -- & -- / 11 / -- & 14 / 14 / 14 & -- / 11 / -- \\
\bottomrule
\end{tabular}%
}
\caption{Options of each evaluation aspect in \textbf{classification tasks} and the number of tasks per language on which they are evaluated (--: unavailable). Where the number of tasks differs between options, the numbers are separated by slashes in the order of the options. Metric and log-probability normalization are given as one row per metric (Accuracy, CondProb, AnsProb) with the three normalizations. Finnish and Norwegian setups cover the multiple-choice tasks with formulation variants (10 of 12 and 14 of 17), while French and Spanish have no formulation variants. \textbf{Generation tasks} (omitted in the table) are compared for the 18 Norwegian tasks only, with both options of greedy decoding and random sampling.}
\label{tab:setup_coverage}
\end{table*}

\section{Appendix: Language-Wise Results}
\label{app:language_results}

\begin{figure*}[t]
\centering
\includegraphics[width=\textwidth]{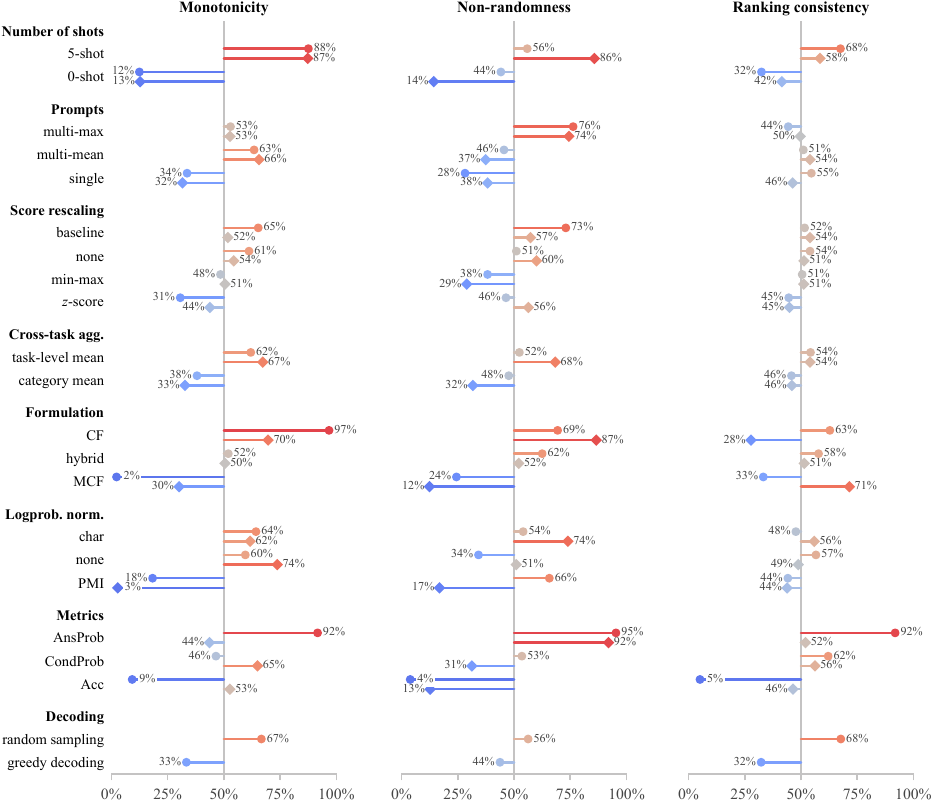}
\caption{Per-option win rates of the 2B models (circle~\redcirc) and larger LLMs (diamond~\reddiam) pooled over languages and task types; win rates are computed for each meta-measure are visualized in the three columns: monotonicity, non-randomness and ranking consistency.}
\label{fig:winrates_2b}
\end{figure*}

\Cref{tab:picks_classification} lists the options selected by the three setup choice methods for both model scales, pooled over languages and for each language. \Cref{tab:ranking_score} gives the option-ranking scores for both model scales. \Cref{fig:winrates_2b} shows the per-option win rates of both model scales under each meta-measure.

\begin{table*}[t]
\centering
\footnotesize
\setlength{\tabcolsep}{1.6pt}
\input{draft_tables/appendix_picks_classification.tex}
\caption{Options selected by the three methods (B: best complete setup, O: option ranking, W: per-option win rate) for classification tasks, pooled over languages and per language, for the 2B models and the larger LLMs. \textbf{Bold}: the method selects a different option than the best complete setup. Grey: the larger LLMs get a different option than the 2B models from the same method. AP: answer probability, CP: conditional answer probability, Acc: accuracy; 5, 0: number of shots; max, mean, sgl: multi-max, multi-mean and single prompt; base, z, mm: baseline, \textit{z}-score and min-max rescaling; mean: task-level mean, cat: category mean; hyb: hybrid; --: no choice (no formulation variants).}
\label{tab:picks_classification}
\end{table*}

\begin{table*}[t]
\centering
\small
\setlength{\tabcolsep}{4pt}
\resizebox{0.4\linewidth}{!}{%
\begin{tabular}{@{}llcc@{}}
\toprule
\textbf{Aspect} & \textbf{Option} & \textbf{2B} & \textbf{Larger} \\
\midrule
Metric & AnsProb & \textbf{0.77} & \textbf{0.61} \\
 & CondProb$^{\dagger}$ & 0.51 & 0.50 \\
 & Acc & 0.38 & 0.46 \\
\addlinespace[2pt]
Norm. & char & \textbf{0.59} & \textbf{0.58} \\
 & PMI & 0.41 & 0.32 \\
 & none & 0.41 & 0.50 \\
\addlinespace[2pt]
Shots & 5-shot & \textbf{0.55} & \textbf{0.61} \\
 & 0-shot & 0.45 & 0.39 \\
\addlinespace[2pt]
Formul.$^{\ddagger}$ & CF & \textbf{0.62} & \textbf{0.52} \\
 & hybrid$^{\dagger}$ & 0.54 & 0.51 \\
 & MCF & 0.36 & 0.47 \\
\addlinespace[2pt]
Prompts & multi-max & \textbf{0.52} & 0.51 \\
 & multi-mean & 0.52 & \textbf{0.52} \\
 & single & 0.47 & 0.47 \\
\addlinespace[2pt]
Task agg. & mean & \textbf{0.51} & \textbf{0.53} \\
 & category mean & 0.50 & 0.48 \\
\addlinespace[2pt]
Rescaling & baseline & \textbf{0.53} & 0.52 \\
 & none & 0.52 & \textbf{0.52} \\
 & min-max & 0.48 & 0.49 \\
 & \textit{z}-score & 0.47 & 0.49 \\
\addlinespace[2pt]
Decoding$^{\dagger}$ & sampling & \textbf{0.56} & -- \\
 & greedy & 0.45 & -- \\
\bottomrule
\end{tabular}%
}
\caption{Option-ranking scores (mean overall score of all setups with the option, over all combinations of the other aspects), averaged over languages, for the 2B models and the
larger LLMs. Options are ordered by the 2B model scores. Bold marks the highest score per aspect and model scale, before rounding into two decimals. $^{\dagger}$Norwegian only. $^{\ddagger}$Finnish and Norwegian.}
\label{tab:ranking_score}
\end{table*}

\begin{comment}
\begin{table}[t]
\centering
\footnotesize
\setlength{\tabcolsep}{4pt}
\input{draft_tables/appendix_picks_generation.tex}
\caption{As Table~\ref{tab:picks_classification}, for the Norwegian generative tasks, on the setups available for both sets of models.}
\label{tab:picks_generation}
\end{table}
\end{comment}

\section{Appendix: When the Meta-Measures Disagree}
\label{app:disagreement}

\begin{table}[H]
\centering
\small
\setlength{\tabcolsep}{8pt}
\resizebox{0.99\linewidth}{!}{%
\begin{tabular}{@{}llll@{}}
\toprule
\textbf{Aspect} & \textbf{Mono} & \textbf{NonRand} & \textbf{RankCons} \\
\midrule
\multicolumn{4}{@{}l}{\textit{Classification tasks}} \\
Formulation & CF 97 & CF 69 & CF 63 \\
Metric & AnsProb 92 & AnsProb 95 & AnsProb 92 \\
Normalization & char 64 & PMI 66 & none 57 \\
Shots & 5-shot 84 & 5-shot 61 & 5-shot 63 \\
Prompts & multi-mean 67 & multi-max 82 & tie (53) \\
Rescaling & baseline 65 & baseline 73 & tie (52) \\
Task agg. & mean 60 & mean 65 & mean 53 \\
\midrule
\multicolumn{4}{@{}l}{\textit{Generative tasks (Norwegian)}} \\
Shots & 5-shot 100 & 0-shot 67 & 5-shot 86 \\
Prompts & multi-max 52 & multi-mean 67 & single 62 \\
Rescaling & none 80 & \textit{z}-score 81 & none 61 \\
Task agg. & mean 71 & category 98 & mean 59 \\
Decoding & sampling 67 & sampling 56 & sampling 68 \\
\bottomrule
\end{tabular}%
}
\caption{Option with the highest win rate (\%) per meta-measure, pooled over languages, for the 2B models.}
\label{tab:disagreement}
\end{table}

The overall score assigns equal weight to the three meta-measures. To examine disagreements among them, we determine for each aspect the option with the highest win rate under each meta-measure separately (\Cref{tab:disagreement}). For the 2B models, the three meta-measures select the same option for formulation, metric, number of shots and task aggregation, i.e., for most of the aspects where all three setup choice methods agree (\Cref{tab:picks_classification}). They disagree on log-probability normalization and prompt aggregation, while ranking consistency provides little separation among rescaling options. Disagreement is more pronounced for Norwegian generative tasks, occurring for four of the five aspects.

\paragraph{Example: maximum versus mean prompt aggregation.}
We compare 672 pairs of setups that differ only in whether scores across prompts are aggregated by the maximum or the mean. Maximum aggregation yields higher non-randomness in 80\% of pairs, whereas mean aggregation yields higher monotonicity in 58\% and slightly higher ranking consistency. The non-randomness advantage of maximum aggregation is driven primarily by its numerator: the gap between the best final score and chance. This gap is larger under maximum aggregation in 94\% of pairs and by 22\% at the median, while the corresponding noise is only 7\% larger. However, this advantage partly results mechanically from taking a maximum. Because scores vary across prompts, the maximum over several prompts exceeds their mean even when the prompts are equally informative, including for a model performing at chance. At the first checkpoint (2B tokens), for example, maximum accuracy already exceeds mean accuracy by 3.0 points on average, compared with a 5.3-point difference at the end of training. Because non-randomness compares this maximum with the chance level of a single prompt, part of the observed advantage reflects the aggregation procedure rather than additional signal. Mean aggregation, by contrast, reduces prompt-level variation and exhibits slightly stronger monotonicity.

\paragraph{Implications.}
Agreement among the meta-measures indicates that a recommendation is insensitive to their relative weighting, as is the case for most high-impact choices. When the meta-measures disagree, the preferred option depends on the desired property. For prompt aggregation, mean aggregation better captures monotonic training progress, whereas maximum aggregation yields greater separation from the single-prompt chance baseline. Such disagreements occur primarily for aspects whose options have similar overall scores (\S\ref{results}), helping explain their relatively low impact on average. More generally, a meta-measure can be inflated by a score transformation without a corresponding increase in useful signal. Maximum aggregation provides one example. Examining individual meta-measures and for non-randomness, its constituent terms can reveal such effects. Finally, disagreements are stronger for the generative tasks. For example, category-mean aggregation wins 98\% of comparisons under non-randomness but only 29\% under monotonicity. These results motivate combining multiple meta-measures rather than relying on any single criterion, while explicitly reporting cases where they disagree.

\end{document}

%% file: draft_tables/appendix_picks_classification.tex
\resizebox{0.99\linewidth}{!}{%
\begin{tabular}{@{}ll@{\hspace{6pt}}ccccccccccccccc@{}}
\toprule
 & & \multicolumn{3}{c}{Pooled} & \multicolumn{3}{c}{Finnish} & \multicolumn{3}{c}{French} & \multicolumn{3}{c}{Norwegian} & \multicolumn{3}{c}{Spanish} \\
\cmidrule(lr){3-5} \cmidrule(lr){6-8} \cmidrule(lr){9-11} \cmidrule(lr){12-14} \cmidrule(lr){15-17}
 & & B & O & W & B & O & W & B & O & W & B & O & W & B & O & W \\
\midrule
Metric & 2B & AP & AP & AP & AP & AP & AP & AP & AP & AP & AP & AP & AP & AP & AP & AP \\
 & Larger & AP & AP & AP & AP & AP & AP & AP & AP & AP & AP & AP & \cellcolor{gray!22}\textbf{Acc} & AP & AP & AP \\
\addlinespace[2pt]
Norm. & 2B & char & char & char & char & char & char & char & char & char & char & char & char & char & char & \textbf{PMI} \\
 & Larger & char & char & char & char & char & \cellcolor{gray!22}\textbf{none} & char & char & char & char & char & char & char & char & \cellcolor{gray!22}char \\
\addlinespace[2pt]
Shots & 2B & 5 & 5 & 5 & 5 & 5 & 5 & 5 & 5 & 5 & 5 & 5 & 5 & 5 & 5 & 5 \\
 & Larger & 5 & 5 & 5 & 5 & 5 & 5 & 5 & 5 & 5 & 5 & 5 & 5 & 5 & 5 & 5 \\
\addlinespace[2pt]
Formul. & 2B & CF & CF & CF & CF & CF & CF & -- & -- & -- & CF & CF & CF & -- & -- & -- \\
 & Larger & CF & CF & CF & CF & CF & CF & -- & -- & -- & \cellcolor{gray!22}hyb & \cellcolor{gray!22}hyb & \cellcolor{gray!22}hyb & -- & -- & -- \\
\addlinespace[2pt]
Prompts & 2B & max & max & max & max & \textbf{mean} & max & max & \textbf{mean} & max & mean & \textbf{max} & \textbf{max} & max & \textbf{sgl} & \textbf{sgl} \\
 & Larger & \cellcolor{gray!22}mean & \cellcolor{gray!22}mean & \textbf{max} & \cellcolor{gray!22}mean & \cellcolor{gray!22}\textbf{max} & \textbf{max} & \cellcolor{gray!22}mean & mean & \textbf{max} & \cellcolor{gray!22}max & max & max & \cellcolor{gray!22}sgl & \cellcolor{gray!22}\textbf{mean} & \cellcolor{gray!22}\textbf{mean} \\
\addlinespace[2pt]
Rescaling & 2B & mm & \textbf{base} & \textbf{base} & mm & \textbf{none} & \textbf{none} & none & none & none & mm & \textbf{base} & \textbf{base} & mm & \textbf{base} & \textbf{base} \\
 & Larger & \cellcolor{gray!22}z & \cellcolor{gray!22}\textbf{none} & \cellcolor{gray!22}\textbf{none} & \cellcolor{gray!22}z & \textbf{none} & \cellcolor{gray!22}z & \cellcolor{gray!22}z & \textbf{none} & \textbf{none} & \cellcolor{gray!22}none & \cellcolor{gray!22}\textbf{mm} & \cellcolor{gray!22}\textbf{mm} & \cellcolor{gray!22}z & \textbf{base} & \textbf{base} \\
\addlinespace[2pt]
Task agg. & 2B & mean & mean & mean & cat & \textbf{mean} & \textbf{mean} & cat & cat & cat & mean & mean & mean & mean & mean & mean \\
 & Larger & mean & mean & mean & \cellcolor{gray!22}mean & mean & mean & cat & \cellcolor{gray!22}\textbf{mean} & \cellcolor{gray!22}\textbf{mean} & mean & mean & mean & mean & mean & mean \\
\bottomrule
\end{tabular}%
}